\documentclass[letterpaper, 10 pt, conference]{ieeeconf}  % Comment this line out if you need a4paper
\IEEEoverridecommandlockouts                              % This command is only needed if 
\usepackage[noadjust]{cite}
\usepackage{graphicx}
\usepackage{amsmath}

\usepackage{multirow}
\usepackage[normalem]{ulem}
\usepackage{hyperref}
\usepackage{algorithm}
\usepackage[noend]{algpseudocode}
\hypersetup{
    colorlinks=true,
    linkcolor=blue,
    filecolor=magenta,      
    urlcolor= blue,
}

\title{\LARGE \bf
Learning to Fly by Crashing
}

\author{
Dhiraj Gandhi, Lerrel Pinto and Abhinav Gupta\\
The Robotics Institute, Carnegie Mellon University\\
\texttt{\{dgandhi, lerrelp, gabhinav\}@andrew.cmu.edu}
}

\begin{document}
\maketitle
\thispagestyle{empty}
\pagestyle{empty}

\begin{abstract}
How do you learn to navigate an Unmanned Aerial Vehicle (UAV) and avoid obstacles? One approach is to use a small dataset collected by human experts: however, high capacity learning algorithms tend to overfit when trained with little data. An alternative is to use simulation. But the gap between simulation and real world remains large especially for perception problems. The reason most research avoids using large-scale real data is the fear of crashes! In this paper, we propose to bite the bullet  and  collect  a  dataset  of  crashes  itself!  We  build  a drone whose sole purpose is to crash into objects: it samples naive trajectories and crashes into random objects. We crash our drone 11,500 times to create one of the biggest UAV crash dataset. This dataset  captures  the different ways in which a  UAV  can crash. We use all this negative flying data in conjunction with positive data  sampled  from  the same  trajectories  to  learn  a  simple  yet  powerful  policy  for  UAV  navigation.  We  show that this simple  self-supervised  model  is  quite  effective  in navigating the UAV even in extremely cluttered environments with dynamic obstacles including humans. For supplementary video see: \url{https://youtu.be/u151hJaGKUo}

\end{abstract}
\section{Introduction}
How do you navigate an autonomous system and avoid obstacles? What is the right approach and data to learn how to navigate? Should we use an end-to-end approach or should there be intermediate representations such as 3D? These are some of the fundamental questions that needs to be answered for solving the indoor navigation of unmanned air vehicle (UAV) and other autonomous systems. Most early research focused on a two-step approach: the first step being perception where either SLAM~\cite{bachrach2009autonomous} or sensors~\cite{zhang2016learning} are used to estimate the underlying map and/or 3D~\cite{bills2011autonomous}; the second step is to use the predicted depth or map to issue motor commands to travel in freespace. While the two step-approach seems reasonable, the cost of sensors and unrecoverable errors from perception make it infeasible. 

Another alternative is to use a monocular camera and learn to predict the motor commands. But how should we learn the mapping from input images to the motor commands? What should be the right data to learn this mapping? One possibility is to use imitation learning~\cite{ross2013learning}. In imitation learning setting, we have a user who provides trajectories to train the flying policy. In order to visit the states not sampled by expert trajectories, they use the learned policy with human corrective actions to train the policy in iterative manner. But learning in such scenarios is restricted to small datasets since human experts are the bottlenecks in providing the training data. Therefore, such approaches cannot exploit high-capacity learning algorithms to train their policies.

\begin{figure}[t!]
\label{fig:drone_intro}
\centering
\includegraphics[width=3.0in]{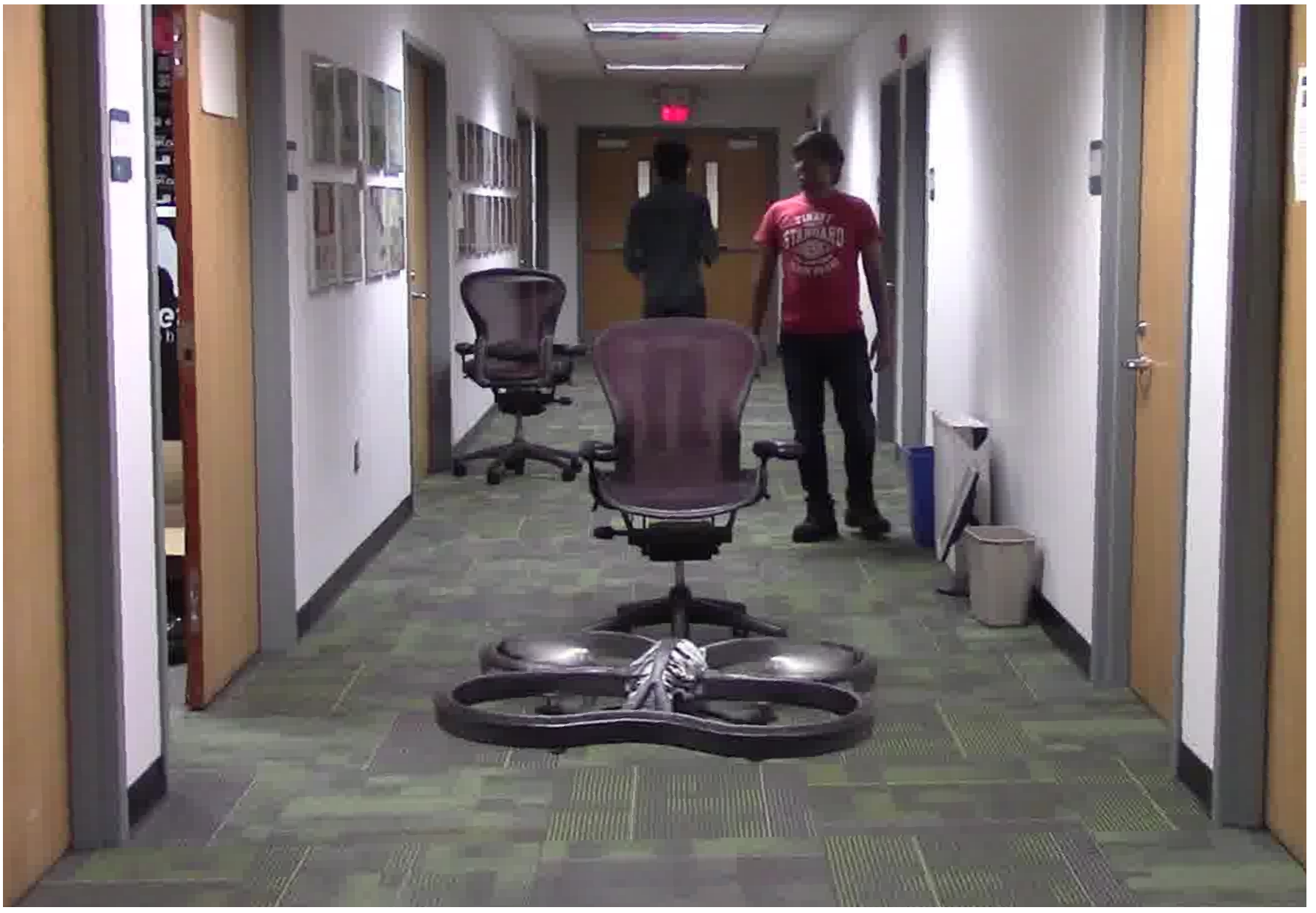}
\caption{In this paper, we focus on learning how to navigate an UAV system. Specifically, we focus on indoor cluttered environment for flying. Instead of using small datasets of imitation learning or performing navigation via intermediate representations such as depth; we collect a large-scale dataset of drone crashes. This dataset acts as a negative instruction and teaches the drone how NOT to crash. We show our simple learning strategy outperforms competitive approaches using the power of large data.
}
\end{figure}

Recently, there has been growing interest in using self-supervised learning for variety of tasks like navigation~\cite{zhu2016target}, grasping~\cite{pinto2016supersizing} and pushing/poking~\cite{agrawal2016learning}. Can we use self-supervised learning to remove the labeling bottleneck of imitation learning? But how do we collect data for self-supervised learning? In contemporary work, Sadeghi and Levine~\cite{DBLP:journals/corr/SadeghiL16} use Reinforcement Learning~(RL) in simulation to train the navigation policy of the drone. They focus on using simulations to avoid collisions that are inevitable since RL-techniques involve a trial-and-error component. They also demonstrate how a policy learned in simulation can transfer to real world without any retraining. But is it really true that simulation-based training can work out of box in real world? Most approaches in computer vision suggest otherwise and require small amounts of real-world data for adaptation. We note that the testing scenarios in ~\cite{DBLP:journals/corr/SadeghiL16} consist mostly of empty corridors where perspective cues are sufficient for navigation~\cite{bills2011autonomous}. These perspective cues are captured in simulation as well. However, when it comes to navigation in cluttered environment (such as one shown in figure~\ref{fig:drone_intro}), the gap between real and simulation widens dramatically.
So, how can we collect data for self-supervised learning in real-world itself? Most approaches avoid self-supervised learning in real-world due to fear of collisions. Instead of finding ways to avoid collisions and crashes, we propose to {\bf bite the bullet} and collect a dataset of crashes itself! We build a drone whose goal is to crash into objects: it samples random trajectories to crash into random objects. We crash our drone {\bf 11,500 times} to create one of the biggest UAV crash dataset. This negative dataset captures the different ways a UAV can crash. It also represents the policy of how UAV should NOT fly. We use all this negative data in conjunction with positive data sampled from the same trajectories to learn a simple yet surprisingly powerful policy for UAV navigation. We show that this simple self-supervised paradigm is quite effective in navigating the UAV even in extremely cluttered environments with dynamic obstacles like humans.
\section{Related work}
This work, which combines self supervised learning with flying a drone in an indoor environment, touches upon the broad fields of robot learning and drone control. We briefly describe these works and their connections to our method.

\subsection{Drone control}
Controlling drones has been a widely studied area motivated by applications in surveillance and transportation. The most prevalent approach is that of SLAM. Several methods~\cite{engel2014lsd,mei2011rslam,bachrach2009autonomous,zhang2012microsoft,henry2012rgb, schmid2013stereo,bills2011autonomous} use range or visual sensors to infer maps of the environment while simultaneously estimating its position in the map. However these SLAM based methods are computationally expensive due to explicit 3D reconstruction, which greatly reduces the ability to have real-time navigation on an inexpensive platform.

Another method, which is closer in real-time applicability to our method, is that of depth estimation methods. One can use onboard range sensors to fly autonomously while avoiding obstacles~\cite{roberts2007quadrotor,bry2012state}. This is however not practical for publically available drones that often have low battery life and low load carrying capacity. One can also use stereo vision based estimation~\cite{achtelik2009stereo,fraundorfer2012vision} with light and cheap cameras. However stereo matching often fails on plain surfaces like the white walls of an office.

Most of the methods described so far use multiple sensors to decide control parameters for the drone. This often leads to higher cost, poor realtime response and bulkier systems. Monocular camera based methods~\cite{bills2011autonomous} use vanishing points as a guidance for drone flying, but still rely on range sensors for collision avoidance. A more recent trend of approaches have been in using learning based methods to infer flying control for the drones. Researchers~\cite{ross2013learning} have used imitation learning strategies to transfer human demonstrations to autonomous navigation. This however fails to collect any negative examples, since humans never crash drones into avoidable obstacles. Because of this data bias, these methods fail to generalize to trajectories outside the training demonstrations from human controllers.

Another recent idea is to use simulators to generate this drone data~\cite{dey_MSR,DBLP:journals/corr/SadeghiL16}. However transferring simulator learned policies to the real world works well only in simplistic scenarios and often require additional training on real-world data.

\subsection{Deep learning for robots}
Learning from trial and error has regained focus in robotics. Self supervised methods~\cite{ levine2016learning, agrawal2016learning, DBLP:journals/corr/PintoG16, pinto2016supersizing}, show how large scale data collection in the real world can be used to learn tasks like grasping and pushing objects in a tabletop environment. Our work extends this idea of self supervision to flying a drone in an indoor environment. Deep reinforcement learning methods~\cite{mnih2015human} have shown impressive results, however they are too data intensive (order of million examples) for our task of drone flying.

A key component of deep learning, is the high amount of data required to train these generalizable models~\cite{krizhevsky2012imagenet,girshick2014rich}. This is where self-supervised learning comes into the picture by allowing the collection of high amounts of data with minimal human supervision. To the best of our knowledge, this is the first large scale effort in collecting more than 40 hours of real drone flight time data which we show is crucial in learning to fly. 
\section{Approach}
We now describe details of our data driven flying approach with discussions on methodologies for data collection and learning. We further describe our hardware setup and implementation for reproducibility. 
\subsection{Hardware Specifications:}
A important goal of our method is to demonstrate the effectiveness of low cost systems for the complex task of flying in an indoor environment. For this purpose, we use the  Parrot Ar-Drone 2.0 which, due to its inaccuracies, is often run in outdoor environments as a hobby drone. We attach no additional sensors/cameras in the flying space in the data collection process. The key required components for the drone is it's inbuilt camera which broadcast 720p resolution images at 30 hz, it's inbuilt accelerometer and a safety hull to collide with objects without damaging the rotors. We externally attach a small camera of the same specification as of  drones inbuilt camera to aid in localization during data collection.

\begin{figure*}[t!]
\centering
\includegraphics[width=7.0in]{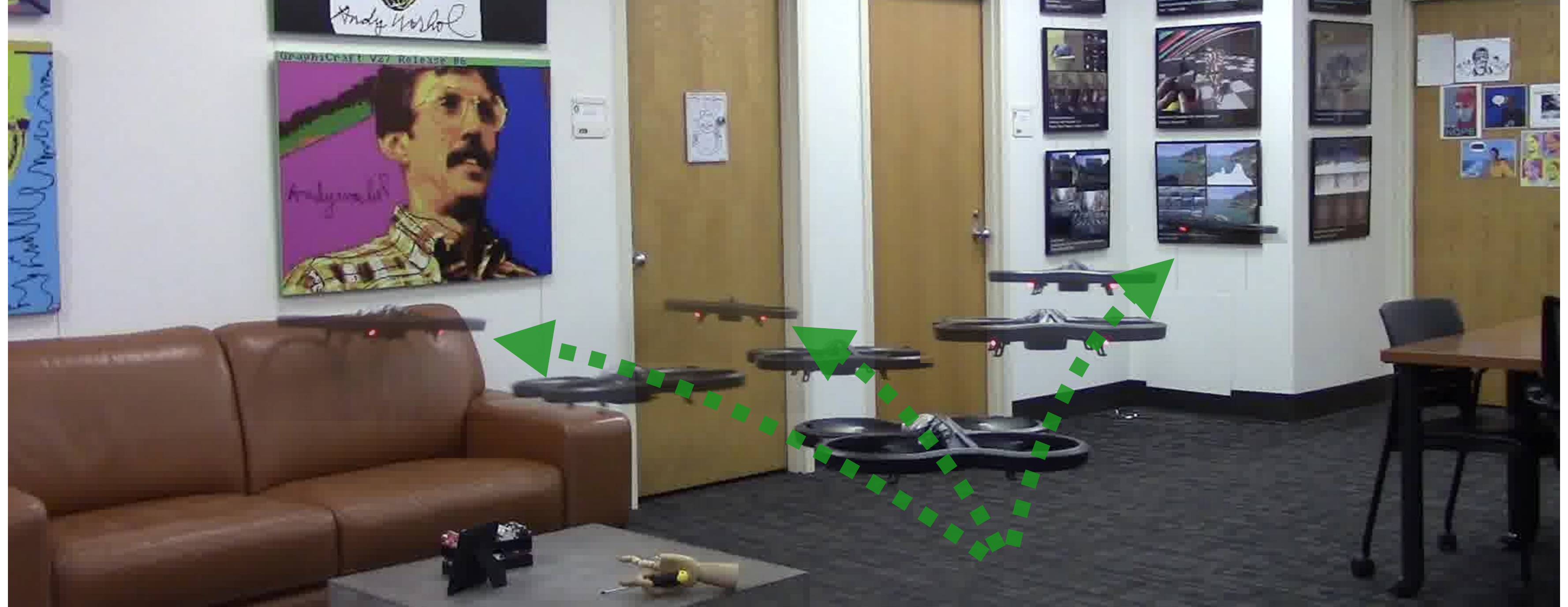}
\caption{We randomly hit objects with our drone more than 11,500 times over a diverse range of environments, This collection of collision trajectories is performed completely autonomously.
}
\label{fig:collecting_traj}
\end{figure*}

\subsection{Data Collection:}
We use a two-step procedure for data collection. First, we sample naive trajectories that lead to collisions with different kind of objects. Based on these sampled we then learn a policy for navigation. These initial naive trajectories and collisions provide a good initialization for reward leading to sampling more complex trajectories. This policy is then used to collect more example akin to hard example mining approach. At this stage, we can obtain much better trajectories that only collide when the learned strategy fails. This sampling of hard negatives has been shown to improve performance ~\cite{pinto2016supersizing,DBLP:journals/corr/ShrivastavaGG16}.

\subsubsection{Collecting collision data}
Most methods for learning with drones~\cite{ross2013learning, bills2011autonomous} often have very few examples of colliding with objects. Our focus is however to collect as much of collision information as possible. This large amount of crashing data should teach our learning model how not to crash/collide with objects. 

But how should we collect this collisions data? One way would be to manually control the drone and crash into objects. But this would severely bias the dataset collected due to human intuition. Another way to collect this data is by commanding the drone to autonomously navigate the environment by SLAM based approaches ~\cite{montemerlo2002fastslam,thrun2004simultaneous} and collect failure cases. But we again have dataset bias issues due to failure modes of SLAM along with sparse collision data. What we need is a method that has low bias to objects it collides with and can also generate lots of collision data. Our proposed method for data collection involves naive random straight line trajectories described in Figure~\ref{fig:collecting_traj}. 

\begin{algorithm}
\caption{Data Collection}\label{data_collection}
\begin{algorithmic}[1]
\State \textbf{Init:} Track position error using PTAM from intial Take Off location
\While{No Crash}
    \State Choose a random direction
    \While {No Collision}
        \State Continue in the same direction
    \EndWhile
    \While {Position error $>$ $\epsilon$ }
        \State calculate control command based on position error
    \EndWhile
\EndWhile

\end{algorithmic}
\end{algorithm}

Algorithm~\ref{data_collection} describes our method of data collection in more detail. The drone is first placed randomly in the environment we desire to collect collision data. The drone then takes off, randomly decides a direction of motion and is commanded to follow a straight line path until collision. After a collision, the drone is commanded to go back to its original position followed by choosing another random direction of motion. This process is continued until the drone cannot recover itself from a collision. After $N_{env\_trial}$ crashes, the drone collision data collection is restarted in a new environment. For each trajectory, we store time stamped images from the camera, estimated trajectories from IMU readings and accelerometer data. The accelerometer data is used to identify the exact moments of collision. Using this, we create our dataset $D=\{d_i\}$ where $d_i=\{I^i_t\}^{N_i}_{t=0}$. For the $i^{th}$ trajectory $d_i$, image $I^i_0$ is the image at the beginning of the trajectory far away from the collision object, while image $I^i_{N_i}$ is the image at the end of the trajectory after $N_i$ timesteps when the drone hits an object.

A key component of this data collection strategy is the ability of the drone to come back to its initial position to collect the next datapoint. However due to cheap low accuracy IMU, it isn't possible to do accurate backtracking using these naive sensors. For this purpose, we use PTAM~\cite{klein07parallel} module that localizes the robot and helps it backtrack.

\begin{figure*}[t!]
\centering
\includegraphics[width=7.0in, height=6.8in]{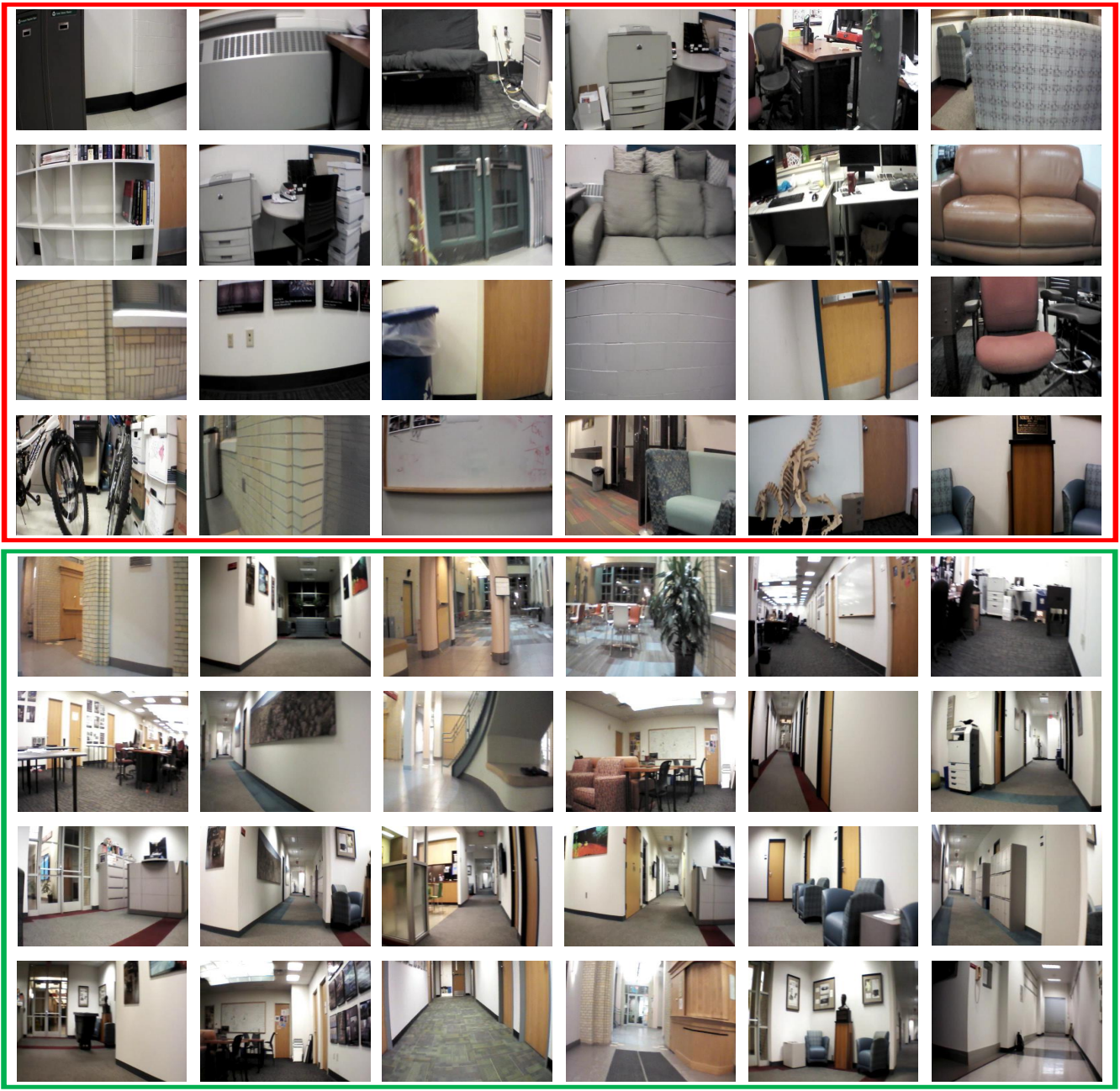}
\caption{Given the data collected from drone collisions, we can extract portions of trajectories into two sections; first one very close to the objects~(red box) \& second far away from the objects~(green box).}
\label{fig:data_collage}
\end{figure*}

We collect 11,500 trajectories of collisions in 20 diverse indoor environments (Figure~\ref{fig:data_collage}). This data is collected over 40 drone flying hours. Note that since the hulls of the drone are cheap and easy to replace, the cost of catastrophic failure is negligible.

\subsubsection{Data processing}
We now describe the annotation procedure for the collected trajectories. The trajectories are first segmented automatically using the accelerometer data. This step restricts each trajectory upto the time of collision. As a next step, we further need to segment the trajectory into positive and negative data i.e. $d_i = d^{+}_i\bigcup d^{-}_i$. Here $d^{+}_i$ is the part of the trajectory far away from the collision object while $d^{-}_i$ is the part of the trajectory close to the colliding object. Note the positive part of the dataset correspond to images where the control signal should be to continue forward. This segmentation is done heuristically by splitting the first $N_{+}$ timesteps of the trajectory as positive and the last $N_{-}$ timesteps as negative trajectories. We ignore the images in the middle part of the trajectory. We now have a dataset with binary classification labels.

\subsection{Learning Methodology}
We now describe our learning methodology given the binary classification dataset we have collected. We will first describe our learning architecture and follow it with description of test time execution.

\subsubsection{Network architecture}
\begin{figure*}[t!]
\centering
\includegraphics[width=6.0in]{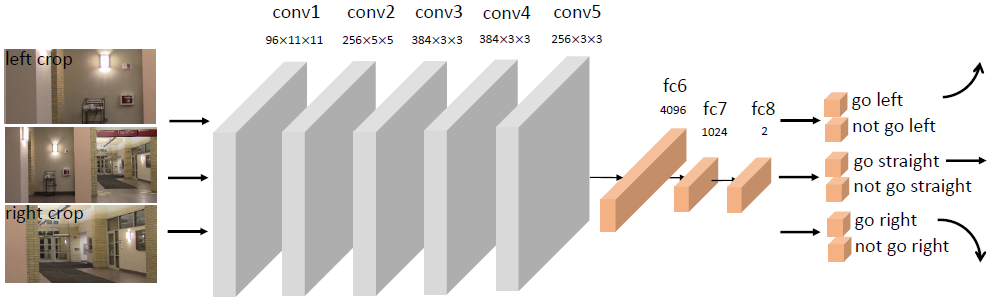}
\caption{We employ deep neural networks to learn how to fly. The convolutional weights of our network (in grey) are pretrained from ImageNet classification~\cite{krizhevsky2012imagenet}, while the fully connected weights (in orange) is initialized randomly and learnt entirely from the collision data. At test-time, crops of the image are given as input, and the network outputs the probability of taking control actions.
}
\label{fig:net}
\end{figure*}

Given the recent successes of deep networks in learning from visual inputs, we employ them to learn controls to fly.  We use the AlexNet architecture~\cite{krizhevsky2012imagenet}. We use ImageNet-pretrained weights as initialization for our network~\cite{girshick2014rich}. This network architecture is represented in Figure~\ref{fig:net}. Note that the weights for last fully connected layer is initialized from a gaussian distribution. We learn a simple classification network which given an input image predicts if the drone should move forward in straight line or not. Therefore, the final layer is a binary softmax and the loss is negative log-likelihood.

\subsubsection{Test time execution}
Our model essentially learns if going straight in a specific direction is good or not. But how can we use this model to fly autonomously in environments with multiple obstacles, narrow corridors and turns? We employ a simple approach to use our binary classification network to fly long distances indoors. Algorithm~\ref{test_fly} succinctly describes this strategy which evaluates the learned network on cropped segments of the drone's image. Based on the right cropped image, complete image and left cropped image network predicts the probability to move in right, straight and left direction. If the straight prediction (P(S)) is greater than  $\alpha$, drone moves forward with the yaw proportional to the difference between the right prediction (P(R)) and left prediction (P(L)). Intuitively, based on the confidence predictions of left and right, we decide to turn the robot left and right while moving forward.
If the prediction for moving straight is below $\alpha$ (going to hit the obstacle soon), we turn the drone left or right depending on which crop of the image predicts move forward.
Intuitively, if the network is fed with only the left part of the image, and the network believes that it is good to go straight there, an optimal strategy would be take a left. In this way we can use a binary classification network to choose more complex directional movements given the image of the scene. This cropping based strategy can also be extended to non planar flying by cropping vertical patches instead of horizontal ones.

\begin{algorithm}
\caption{Policy for flying indoor}
\label{test_fly}
\begin{algorithmic}[1]
\While{No Crash}
    \State \textbf{Input : } Real time image
    \State{P(L), P(S), P(R)} = Network\{image, left \&\ right crop\}
    \If {P(S) $>$ $\alpha$ }
        \State Linear Velocity = $\beta$
        \State Angular Velocity $\propto$ P(R) - P(L)
    \Else
        \State Linear Velocity = 0
        \If{P(R) $>$ P(L)}
            \While{P(S) $<$ $\alpha$}
                \State P(S) = Network\{image\}
                \State Take Right turn
        \EndWhile
        \Else
            \While{P(S) $<$ $\alpha$}
            \State P(S) = Network\{image\}
            \State Take Left turn
        \EndWhile
        \EndIf 
    \EndIf
\EndWhile

\end{algorithmic}
\end{algorithm}

\section{Experimental Evaluation}
We now describe the evaluation procedure of our method on indoor environments and compare with strong depth driven baselines. 

\begin{figure}[h!]
 \centering
 \includegraphics[width=3.2in]{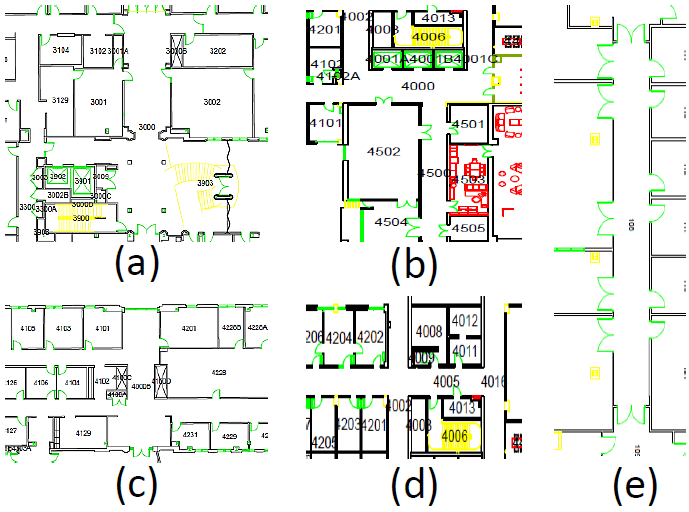}
 \caption{Floor plans for the testing environment are shown here: (a) NSH Entrance, (b) Wean Hall, (c) NSH $4^{th}$ Floor, (d) Glass Door and (e) Hallway. Note that `Hallway with Chairs' environment has the same floorplan as `Hallway' but with chairs as additional obstacles.
 }
 \label{fig:floorplan}
 \end{figure} 
 
\begin{figure*}[t!]
\centering
\includegraphics[width=6.5in]{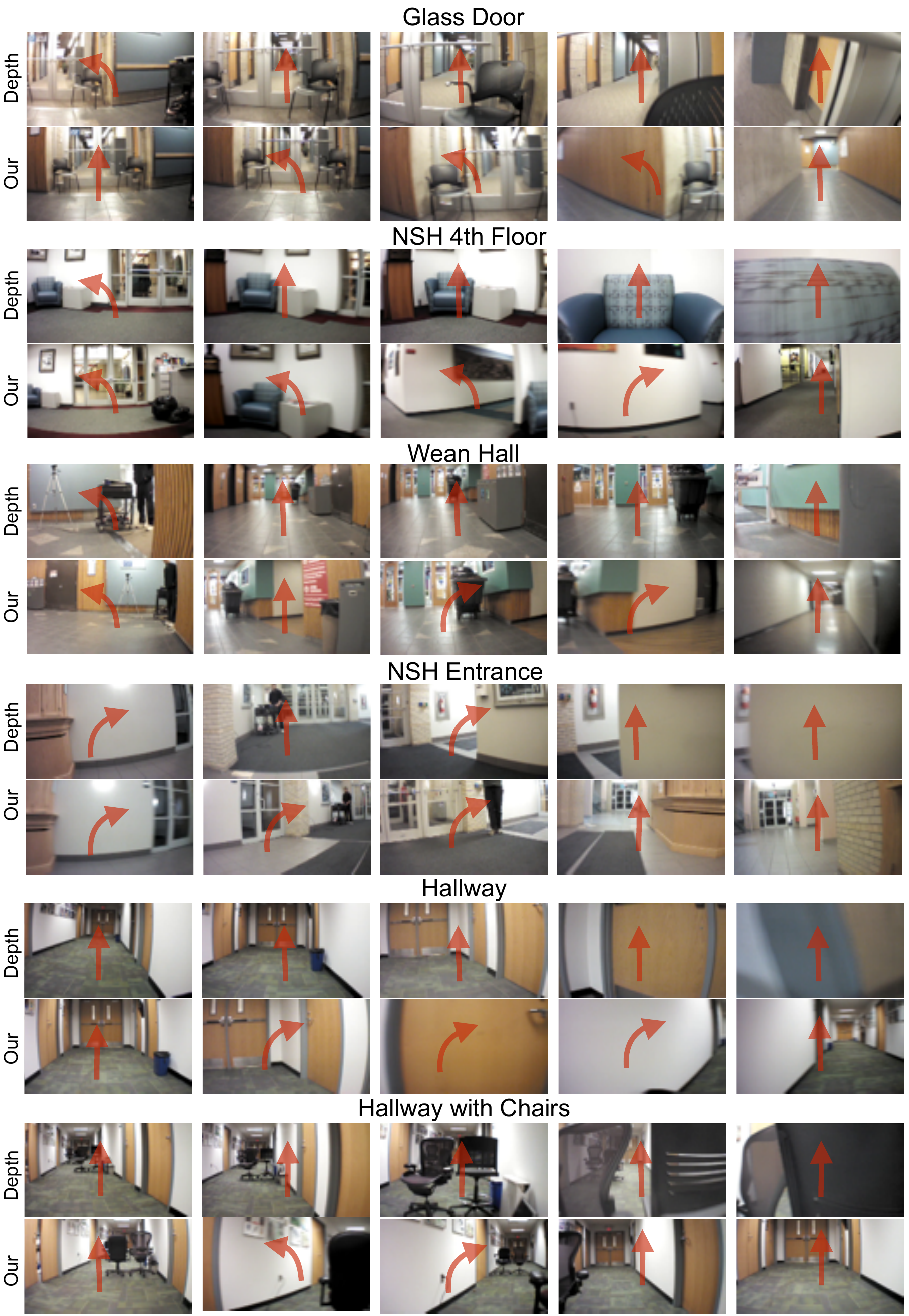}
\caption{Here we show the comparisons of the trajectories of our method vs the strong baseline of depth based prediction on our testing environments. The arrows denote the action taken by corresponding method. 
}
\label{fig:results}
\end{figure*}

\begin{table*}[]
\centering
\caption{Average distance and average time before collision}
\label{tab:res}
\begin{tabular}{rcccccc}
\multicolumn{1}{l|}{} & \multicolumn{2}{c|}{Glass door} & \multicolumn{2}{c|}{NSH 4th Floor} & \multicolumn{2}{c}{NSH Entrance } \\ \cline{2-7} 
\multicolumn{1}{l|}{} & \multicolumn{1}{c|}{Avg Dist (m)} & \multicolumn{1}{c|}{Avg Time (s)} & \multicolumn{1}{c|}{Avg Dist (m)} & \multicolumn{1}{c|}{Avg Time (s)} & \multicolumn{1}{c|}{Avg Dist (m)} & \multicolumn{1}{c}{Avg Time (s)} \\ \hline
\multicolumn{1}{r|}{Best Straight} & 3.3 & \multicolumn{1}{c|}{3.0} & 6.2 & \multicolumn{1}{c|}{7.0} & 2.7 & 3.0 \\
\multicolumn{1}{r|}{Depth Prediction} & 3.1 & \multicolumn{1}{c|}{5.0} & 14.0 & \multicolumn{1}{c|}{28.3} & 13.4 & 22.6 \\
\multicolumn{1}{r|}{Our Method} & 27.9 & \multicolumn{1}{c|}{56.6} & 54.0 & \multicolumn{1}{c|}{120.4} & 42.3 & 78.4 \\
\multicolumn{1}{r|}{Human} & 84.0 & \multicolumn{1}{c|}{145.0} & 99.9 & \multicolumn{1}{c|}{209.0} & 119.6 & 196.0 \\
\multicolumn{1}{l}{} & \multicolumn{1}{l}{} & \multicolumn{1}{l}{} & \multicolumn{1}{l}{} & \multicolumn{1}{l}{} & \multicolumn{1}{l}{} & \multicolumn{1}{l}{} \\
\multicolumn{1}{l|}{} & \multicolumn{2}{c|}{Hallway} & \multicolumn{2}{c|}{Hallway With Chairs} & \multicolumn{2}{c}{Wean Hall} \\ \cline{2-7} 
\multicolumn{1}{l|}{} & \multicolumn{1}{c|}{Avg Dist (m)} & \multicolumn{1}{c|}{Avg Time (s)} & \multicolumn{1}{c|}{Avg Dist (m)} & \multicolumn{1}{c|}{Avg Time (s)} & \multicolumn{1}{c|}{Avg Dist (m)} & \multicolumn{1}{c}{Avg Time (s)} \\ \hline
\multicolumn{1}{r|}{Best straight} & 6.2 & \multicolumn{1}{c|}{6.6} & 2.7 & \multicolumn{1}{c|}{3.1} & 4.2 & 4.6 \\
\multicolumn{1}{r|}{Depth Prediction} & 24.9 & \multicolumn{1}{c|}{26.6} & 25.5 & \multicolumn{1}{c|}{43.5} & 11.6 & 22.1 \\
\multicolumn{1}{r|}{Our Method} & 115.2 & \multicolumn{1}{c|}{210.0} & 86.9 & \multicolumn{1}{c|}{203.5} & 22.4 & 47.0 \\
\multicolumn{1}{r|}{Human} & 95.7 & \multicolumn{1}{c|}{141.0} & 69.3 & \multicolumn{1}{c|}{121.0} & 70.6 & 126.0
\end{tabular}
\end{table*}

\subsection{Baselines}
To evaluate our model, we compare the performance with a Straight line policy, a Depth prediction based policy and a human controlled policy.

\subsubsection{Straight line policy}
A weak baseline for indoor navigation is to take an open-loop straight line path. For environments that contain narrow corridors, this will fail since errors compound which results in curved paths that will hit the walls. To strengthen this baseline, we choose the best (oracle) direction for the straight line policy.

\subsubsection{Depth prediction based policy}
Recent advances in depth estimation from monocular cameras~\cite{eigen2014depth} have shown impressive results. These models for depth estimation are often trained over around 220k indoor depth data. A strong baseline is to use this depth prediction network to generate a depth maps given a monocular image and make an inference on the direction of movement based on this depth map.

\subsubsection{Human policy}
One of the strongest baseline is to use a human operator to fly the drone. In this case, we ask participants to control the drone only given the monocular image seen by the drone. The participants then use a joystick to give the commanded direction of motion to the drone. We also allow the participants to fly the drone in a test trial so that they get familiar with the response of the drone.

\subsection{Testing Environments}
To show the generalizaility of our method, we test it on 6 complex indoor environments: `Glass Door', `NSH $4^{th}$ Floor', `NSH Entrance', `Wean Hall', `Hallway' and `Hallway with Chairs'. Floor plans for these environments can be seen in Figure~\ref{fig:floorplan}. These environments have unique challenges that encompas most of the challenges faced in general purpose indoor navigation. For each of these environments, our method along with all the baselines are run 5 times with different start orientations and positions. This is done to ensure that the comparisons are robust to initializations.

\subsubsection{Glass Door}
 This environment is corridor with a corner that has transparent doors. The challenge for the drone is to take a turn at this corner without crashing into the transparent door. Most depth based estimation techniques are prone to fail in this scenario (depth sensing is poor with transparent objects). However data driven visual techniques have been shown to perform reasonably with these tough transparent obstacles.
 
\subsubsection{NSH $4^{th}$ Floor}
The second environment we test on is the office space of the $4^{th}$ floor of the Newell Simon Hall at Carnegie Mellon University. In this environment, the drone has the capacity to navigate through a larger environment with several turns in narrow corridors.

\subsubsection{NSH Entrance}
In this environment, the drone is initialized at the entrance of the Newell Simon Hall at Carnegie Mellon University. Here the challenge is manoeuvre through a hall with glass walls and into an open space (atrium) that is cluttered with dining tables and chairs. 

\subsubsection{Wean Hall}
This environment has small stretches of straight path which are connected at 90 degree to each other. Drone needs to take  required turn to avoid collision at the intersection.

\subsubsection{Hallway}
This is a narrow~($2m$) dead end straight corridor where drone has keep in the center to avoid the collision with the wall. At the dead end, the drone needs to take a turn and fly back. While flying back, the drone will again meet the dead end and can turn back again. This environment tests the long term flight capacity of the drone.

\subsubsection{Hallway With chairs}
To the explicitly test the robustness of the controller to clutter and obstacles, we modify `Smith Hallway' with chairs. The controller has to identify the narrow gaps between the chairs and the walls to avoid collisions. This gap can be less than 1m at some points.

We would like to point out that 2 out of 6 environments were also seen during training. These correspond to the NSH 4th Floor and NSH Entrance. The remaining 4 environments are completely novel.

\subsection{Results}
To evaluate the performance of different baselines, we used average distance and average time of flight without collisions as the metric of evaluation. This metric also terminates flight runs when they take small loops (spinning on spot). Quantitative results are presented in Table~\ref{tab:res}. We also qualitatively show in Figure~\ref{fig:results} the comparison of trajectories generated by our method vs the depth prediction based baseline.

On every environment/setting we test on, we see that our method performs much better than the depth baseline. The best straight baseline provides an estimate of how difficult the environments are. The human controlled baselines are higher than our method for most environments. However for some environments like `Hallway with Chairs', the presence of cluttered objects makes it difficult for the participants to navigate through narrow spaces which allows our method to surpass human level control in this environment. 

A key observation is the failure of depth based methods to (a) glass walls and doors, and (b) untextured flat walls. Glass walls give the depth based models the impression that the closest obstacle is farther away. However our method, since it has been seen examples of collisions with glass windows may have latched onto features that help it avoid glass walls or doors. Another difficulty depth based models face are in untextured environments like corridors. Since our model has already seen untextured environments during training, it has learned to identify corners and realise that moving towards these corridor corners isn't desirable.

The results on the `Hallway' environments further cements our method's claim by autonomously navigating for more than 3 minutes (battery life of drone in flight use is 5 minutes).

\section{Conclusion}
We propose a data driven approach to learn to fly by crashing more than 11,500 times in a multitude of diverse training environments. These crashing trajectories generate the biggest (to our knowledge) UAV crashing dataset and demonstrates the importance of negative data in learning. A standard deep network architecture is trained on this indoor crashing data, with the task of binary classification. By learning how to NOT fly, we show that even simple strategies easily outperforms depth prediction based methods on a variety of testing environments and is comparable to human control on some environments. This work demonstrates: (a) it is possible to collect self-supervised data for navigation at large-scale; (b) such data is crucial for learning how to navigate.

\section*{ACKNOWLEDGEMENTS}
This work was supported by ONR MURI N000141612007, NSF IIS-1320083 and Google Focused Award. AG was supported in part by Sloan Research Fellowship.

\bibliographystyle{IEEEtran}
\bibliography{IEEEabrv,references}

\begin{thebibliography}{10}
\providecommand{\url}[1]{#1}
\csname url@samestyle\endcsname
\providecommand{\newblock}{\relax}
\providecommand{\bibinfo}[2]{#2}
\providecommand{\BIBentrySTDinterwordspacing}{\spaceskip=0pt\relax}
\providecommand{\BIBentryALTinterwordstretchfactor}{4}
\providecommand{\BIBentryALTinterwordspacing}{\spaceskip=\fontdimen2\font plus
\BIBentryALTinterwordstretchfactor\fontdimen3\font minus
  \fontdimen4\font\relax}
\providecommand{\BIBforeignlanguage}[2]{{%
\expandafter\ifx\csname l@#1\endcsname\relax
\typeout{** WARNING: IEEEtran.bst: No hyphenation pattern has been}%
\typeout{** loaded for the language `#1'. Using the pattern for}%
\typeout{** the default language instead.}%
\else
\language=\csname l@#1\endcsname
\fi
#2}}
\providecommand{\BIBdecl}{\relax}
\BIBdecl

\bibitem{bachrach2009autonomous}
A.~G. Bachrach, ``Autonomous flight in unstructured and unknown indoor
  environments,'' Ph.D. dissertation, Massachusetts Institute of Technology,
  2009.

\bibitem{zhang2016learning}
T.~Zhang, G.~Kahn, S.~Levine, and P.~Abbeel, ``Learning deep control policies
  for autonomous aerial vehicles with mpc-guided policy search,'' in
  \emph{Robotics and Automation (ICRA), 2016 IEEE International Conference
  on}.\hskip 1em plus 0.5em minus 0.4em\relax IEEE, 2016, pp. 528--535.

\bibitem{bills2011autonomous}
C.~Bills, J.~Chen, and A.~Saxena, ``Autonomous mav flight in indoor
  environments using single image perspective cues,'' in \emph{Robotics and
  automation (ICRA), 2011 IEEE international conference on}.\hskip 1em plus
  0.5em minus 0.4em\relax IEEE, 2011, pp. 5776--5783.

\bibitem{ross2013learning}
S.~Ross, N.~Melik-Barkhudarov, K.~S. Shankar, A.~Wendel, D.~Dey, J.~A. Bagnell,
  and M.~Hebert, ``Learning monocular reactive uav control in cluttered natural
  environments,'' in \emph{Robotics and Automation (ICRA), 2013 IEEE
  International Conference on}.\hskip 1em plus 0.5em minus 0.4em\relax IEEE,
  2013, pp. 1765--1772.

\bibitem{zhu2016target}
Y.~Zhu, R.~Mottaghi, E.~Kolve, J.~J. Lim, A.~Gupta, L.~Fei-Fei, and A.~Farhadi,
  ``Target-driven visual navigation in indoor scenes using deep reinforcement
  learning,'' \emph{arXiv preprint arXiv:1609.05143}, 2016.

\bibitem{pinto2016supersizing}
L.~Pinto and A.~Gupta, ``Supersizing self-supervision: Learning to grasp from
  50k tries and 700 robot hours,'' \emph{ICRA}, 2016.

\bibitem{agrawal2016learning}
P.~Agrawal, A.~Nair, P.~Abbeel, J.~Malik, and S.~Levine, ``Learning to poke by
  poking: Experiential learning of intuitive physics,'' \emph{arXiv preprint
  arXiv:1606.07419}, 2016.

\bibitem{DBLP:journals/corr/SadeghiL16}
\BIBentryALTinterwordspacing
F.~Sadeghi and S.~Levine, ``(cad){\textdollar}{\^{}}2{\textdollar}rl: Real
  single-image flight without a single real image,'' \emph{CoRR}, vol.
  abs/1611.04201, 2016. [Online]. Available:
  \url{http://arxiv.org/abs/1611.04201}
\BIBentrySTDinterwordspacing

\bibitem{engel2014lsd}
J.~Engel, T.~Sch{\"o}ps, and D.~Cremers, ``Lsd-slam: Large-scale direct
  monocular slam,'' in \emph{European Conference on Computer Vision}.\hskip 1em
  plus 0.5em minus 0.4em\relax Springer, 2014, pp. 834--849.

\bibitem{mei2011rslam}
C.~Mei, G.~Sibley, M.~Cummins, P.~Newman, and I.~Reid, ``Rslam: A system for
  large-scale mapping in constant-time using stereo,'' \emph{International
  journal of computer vision}, vol.~94, no.~2, pp. 198--214, 2011.

\bibitem{zhang2012microsoft}
Z.~Zhang, ``Microsoft kinect sensor and its effect,'' \emph{IEEE multimedia},
  vol.~19, no.~2, pp. 4--10, 2012.

\bibitem{henry2012rgb}
P.~Henry, M.~Krainin, E.~Herbst, X.~Ren, and D.~Fox, ``Rgb-d mapping: Using
  kinect-style depth cameras for dense 3d modeling of indoor environments,''
  \emph{The International Journal of Robotics Research}, vol.~31, no.~5, pp.
  647--663, 2012.

\bibitem{schmid2013stereo}
K.~Schmid, T.~Tomic, F.~Ruess, H.~Hirschm{\"u}ller, and M.~Suppa, ``Stereo
  vision based indoor/outdoor navigation for flying robots,'' in
  \emph{Intelligent Robots and Systems (IROS), 2013 IEEE/RSJ International
  Conference on}.\hskip 1em plus 0.5em minus 0.4em\relax IEEE, 2013, pp.
  3955--3962.

\bibitem{roberts2007quadrotor}
J.~F. Roberts, T.~Stirling, J.-C. Zufferey, and D.~Floreano, ``Quadrotor using
  minimal sensing for autonomous indoor flight,'' in \emph{European Micro Air
  Vehicle Conference and Flight Competition (EMAV2007)}, no. LIS-CONF-2007-006,
  2007.

\bibitem{bry2012state}
A.~Bry, A.~Bachrach, and N.~Roy, ``State estimation for aggressive flight in
  gps-denied environments using onboard sensing,'' in \emph{Robotics and
  Automation (ICRA), 2012 IEEE International Conference on}.\hskip 1em plus
  0.5em minus 0.4em\relax IEEE, 2012, pp. 1--8.

\bibitem{achtelik2009stereo}
M.~Achtelik, A.~Bachrach, R.~He, S.~Prentice, and N.~Roy, ``Stereo vision and
  laser odometry for autonomous helicopters in gps-denied indoor
  environments,'' in \emph{SPIE Defense, security, and sensing}.\hskip 1em plus
  0.5em minus 0.4em\relax International Society for Optics and Photonics, 2009,
  pp. 733\,219--733\,219.

\bibitem{fraundorfer2012vision}
F.~Fraundorfer, L.~Heng, D.~Honegger, G.~H. Lee, L.~Meier, P.~Tanskanen, and
  M.~Pollefeys, ``Vision-based autonomous mapping and exploration using a
  quadrotor mav,'' in \emph{Intelligent Robots and Systems (IROS), 2012
  IEEE/RSJ International Conference on}.\hskip 1em plus 0.5em minus 0.4em\relax
  IEEE, 2012, pp. 4557--4564.

\bibitem{dey_MSR}
S.~Shah, D.~Dey, C.~Lovett, and A.~Kapoor, ``{A}erial {I}nformatics and
  {R}obotics platform,'' Microsoft Research, Tech. Rep.
  {M}{S}{R}-{T}{R}-2017-9, 2017.

\bibitem{levine2016learning}
S.~Levine, P.~Pastor, A.~Krizhevsky, and D.~Quillen, ``Learning hand-eye
  coordination for robotic grasping with deep learning and large-scale data
  collection,'' \emph{ISER}, 2016.

\bibitem{DBLP:journals/corr/PintoG16}
\BIBentryALTinterwordspacing
L.~Pinto and A.~Gupta, ``Learning to push by grasping: Using multiple tasks for
  effective learning,'' \emph{CoRR}, vol. abs/1609.09025, 2016. [Online].
  Available: \url{http://arxiv.org/abs/1609.09025}
\BIBentrySTDinterwordspacing

\bibitem{mnih2015human}
V.~Mnih, K.~Kavukcuoglu, D.~Silver, A.~A. Rusu, J.~Veness, M.~G. Bellemare,
  A.~Graves, M.~Riedmiller, A.~K. Fidjeland, G.~Ostrovski \emph{et~al.},
  ``Human-level control through deep reinforcement learning,'' \emph{Nature},
  2015.

\bibitem{krizhevsky2012imagenet}
A.~Krizhevsky, I.~Sutskever, and G.~E. Hinton, ``Imagenet classification with
  deep convolutional neural networks,'' in \emph{NIPS}, 2012.

\bibitem{girshick2014rich}
R.~Girshick, J.~Donahue, T.~Darrell, and J.~Malik, ``Rich feature hierarchies
  for accurate object detection and semantic segmentation,'' in
  \emph{Proceedings of the IEEE conference on computer vision and pattern
  recognition}, 2014, pp. 580--587.

\bibitem{DBLP:journals/corr/ShrivastavaGG16}
\BIBentryALTinterwordspacing
A.~Shrivastava, A.~Gupta, and R.~B. Girshick, ``Training region-based object
  detectors with online hard example mining,'' \emph{CoRR}, vol.
  abs/1604.03540, 2016. [Online]. Available:
  \url{http://arxiv.org/abs/1604.03540}
\BIBentrySTDinterwordspacing

\bibitem{montemerlo2002fastslam}
M.~Montemerlo, S.~Thrun, D.~Koller, B.~Wegbreit \emph{et~al.}, ``Fastslam: A
  factored solution to the simultaneous localization and mapping problem,''
  2002.

\bibitem{thrun2004simultaneous}
S.~Thrun, Y.~Liu, D.~Koller, A.~Y. Ng, Z.~Ghahramani, and H.~Durrant-Whyte,
  ``Simultaneous localization and mapping with sparse extended information
  filters,'' \emph{The International Journal of Robotics Research}, vol.~23,
  no. 7-8, pp. 693--716, 2004.

\bibitem{klein07parallel}
G.~Klein and D.~Murray, ``Parallel tracking and mapping for small {AR}
  workspaces,'' in \emph{Proc. Sixth {IEEE} and {ACM} International Symposium
  on Mixed and Augmented Reality {(ISMAR'07)}}, Nara, Japan, November 2007.

\bibitem{eigen2014depth}
D.~Eigen, C.~Puhrsch, and R.~Fergus, ``Depth map prediction from a single image
  using a multi-scale deep network,'' in \emph{Advances in neural information
  processing systems}, 2014, pp. 2366--2374.

\end{thebibliography}
\end{document}